\documentclass[10pt,twocolumn,letterpaper]{article}

\usepackage{cvpr}              % To produce the CAMERA-READY version
\definecolor{cvprblue}{rgb}{0.21,0.49,0.74}
\usepackage[pagebackref,colorlinks,allcolors=cvprblue]{hyperref}
\usepackage{arydshln}
\usepackage{multirow}
\usepackage{algorithm}
\usepackage{algorithmic}

\def\paperID{xxxx} % *** Enter the Paper ID here
\def\confName{CVPR}
\def\confYear{2025}

\title{SegAnyPET: Universal Promptable Segmentation from Positron Emission Tomography Images}

\author{Yichi Zhang\textsuperscript{1,2,*} \  Le Xue\textsuperscript{1,2,*} \  Wenbo Zhang\textsuperscript{1,2} \ Lanlan Li\textsuperscript{1} \ Yuchen Liu\textsuperscript{1,2} \ Chen Jiang\textsuperscript{2} \ \\ Yuan Cheng\textsuperscript{1,2,\dag} \ Yuan Qi\textsuperscript{1,2,\dag}  \\ [2.5mm]
\textsuperscript{1}Fudan University \
\textsuperscript{2}Shanghai Academy of Artificial Intelligence for Science}

\begin{document}
\maketitle
\begin{abstract}
Positron Emission Tomography (PET) imaging plays a crucial role in modern medical diagnostics by revealing the metabolic processes within a patient's body, which is essential for quantification of therapy response and monitoring treatment progress. However, the segmentation of PET images presents unique challenges due to their lower contrast and less distinct boundaries compared to other structural medical modalities. Recent developments in segmentation foundation models have shown superior versatility across diverse natural image segmentation tasks. Despite the efforts of medical adaptations, these works primarily focus on structural medical images with detailed physiological structural information and exhibit poor generalization ability when adapted to molecular PET imaging.
In this paper, we collect and construct PETS-5k, the largest PET segmentation dataset to date, comprising 5,731 three-dimensional whole-body PET images and encompassing over 1.3M 2D images. Based on the established dataset, we develop SegAnyPET, a modality-specific 3D foundation model for universal promptable segmentation from PET images. To issue the challenge of discrepant annotation quality of PET images, we adopt a cross prompting confident learning (CPCL) strategy with an uncertainty-guided self-rectification process to robustly learn segmentation from high-quality labeled data and low-quality noisy labeled data. Experimental results demonstrate that SegAnyPET can correctly segment seen and unseen targets using only one or a few prompt points, outperforming state-of-the-art foundation models and task-specific fully supervised models with higher accuracy and strong generalization ability for universal segmentation. As the first foundation model for PET images, we believe that SegAnyPET will advance the applications to various downstream tasks for molecular imaging.\footnote{The model and code will be publicly available at \href{https://github.com/YichiZhang98/SegAnyPET}{the project page}. \\ * equal contribution. \dag   corresponding authors.}
\end{abstract}    
\section{Introduction}
\label{sec:intro}

\begin{figure}[t]
	\includegraphics[width=\linewidth]{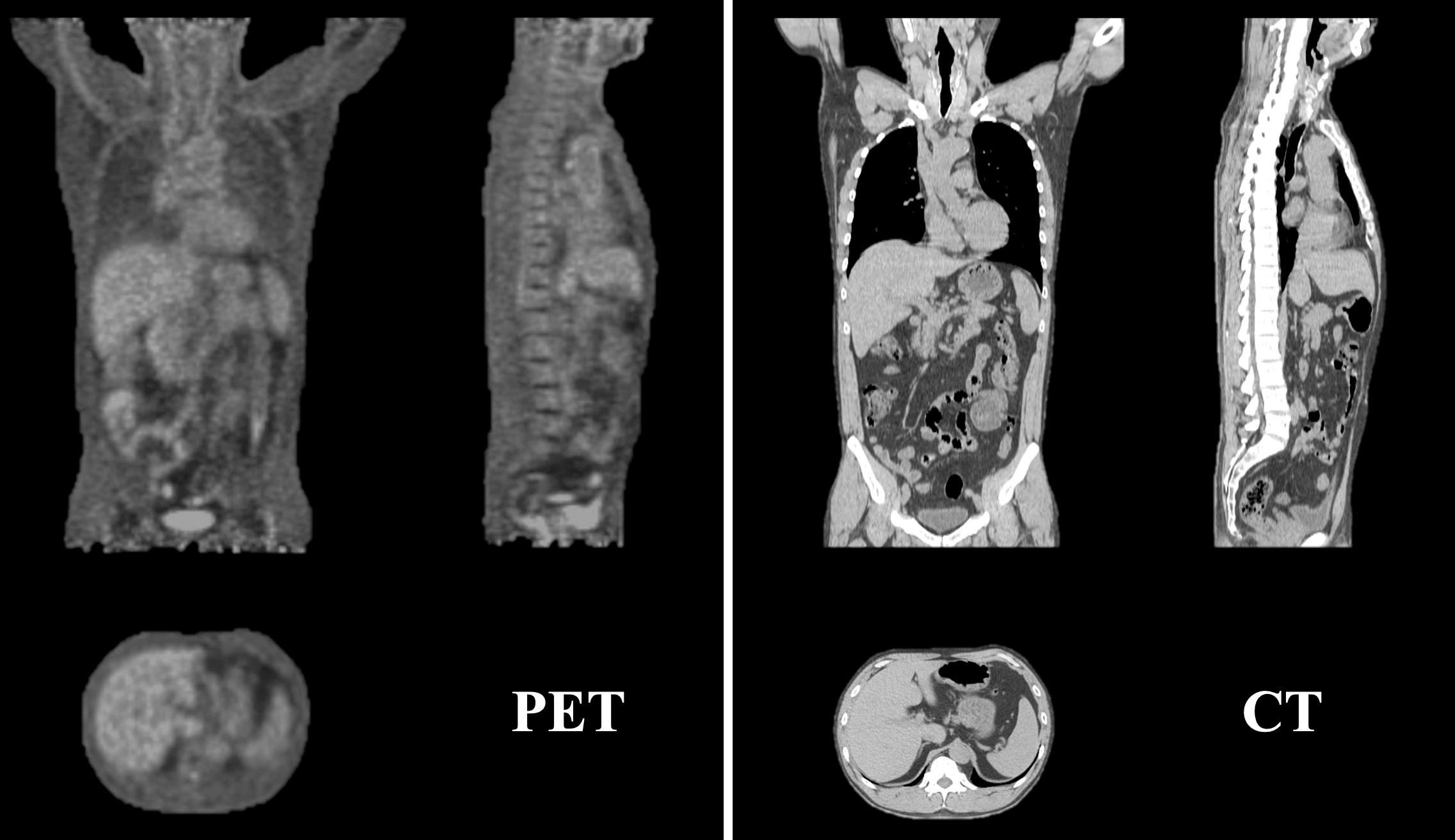}
	\caption{Visual comparison of an example case between molecular PET image and structural CT image.}
	\label{Visualize}
\end{figure}

%-------------------------------------------------------------------------
Medical imaging technology has become an indispensable tool in modern clinical diagnosis \cite{kasban2015comparative,Lynch2018NewMT}. The steady development of increasingly sophisticated imaging techniques has led to a dramatic increase in patient treatment outcomes, especially when diseases can only be observed internally within the patient's body \cite{MIA2017survey}. 
Segmentation of targets such as organs and tumors from medical images is one of the most representative and comprehensive research topics in both the computer vision and medical image analysis communities \cite{hesamian2019deep}. 
Accurate segmentation can provide reliable volumetric and shape information on target structures and assist in many clinical applications \cite{lalande2021deep,AbdomenCT-1K,zhang2024nasalseg}.

In addition to commonly used medical modalities such as computed tomography (CT) and magnetic resonance imaging (MRI), which can provide detailed physiological structural information, positron emission tomography (PET) has also been widely applied in medical examinations as an emerging molecular imaging modality \cite{schwenck2023advances}.
The purpose of PET imaging is to reveal the ongoing metabolic processes within the body of a patient. A radioactive tracer is injected into the patient, typically containing the radioactive isotope fluorine-18 synthesized fluorodeoxyglucose ($^{18}$F-FDG). 
The radioactive decay emits positrons, leading to gamma photons, which are produced by annihilation and captured by detectors surrounding the patient to create a 3D map of metabolic activity.
In FDG-PET examinations, the tracer is used to assess local glucose uptake and evaluate organ metabolism \cite{ren2019atlas,wang2025robust} and presence of tumor metastasis \cite{AutoPET}, so as to monitor progress during the treatment process.
However, due to the low resolution nature and the partial volume effects, PET images exhibit lower image quality and have less distinct boundaries compared to other structural images such as CT scans (see \cref{Visualize}), which presents a significant challenge for accurate segmentation.

\begin{figure*}[t]
	\includegraphics[width=\linewidth]{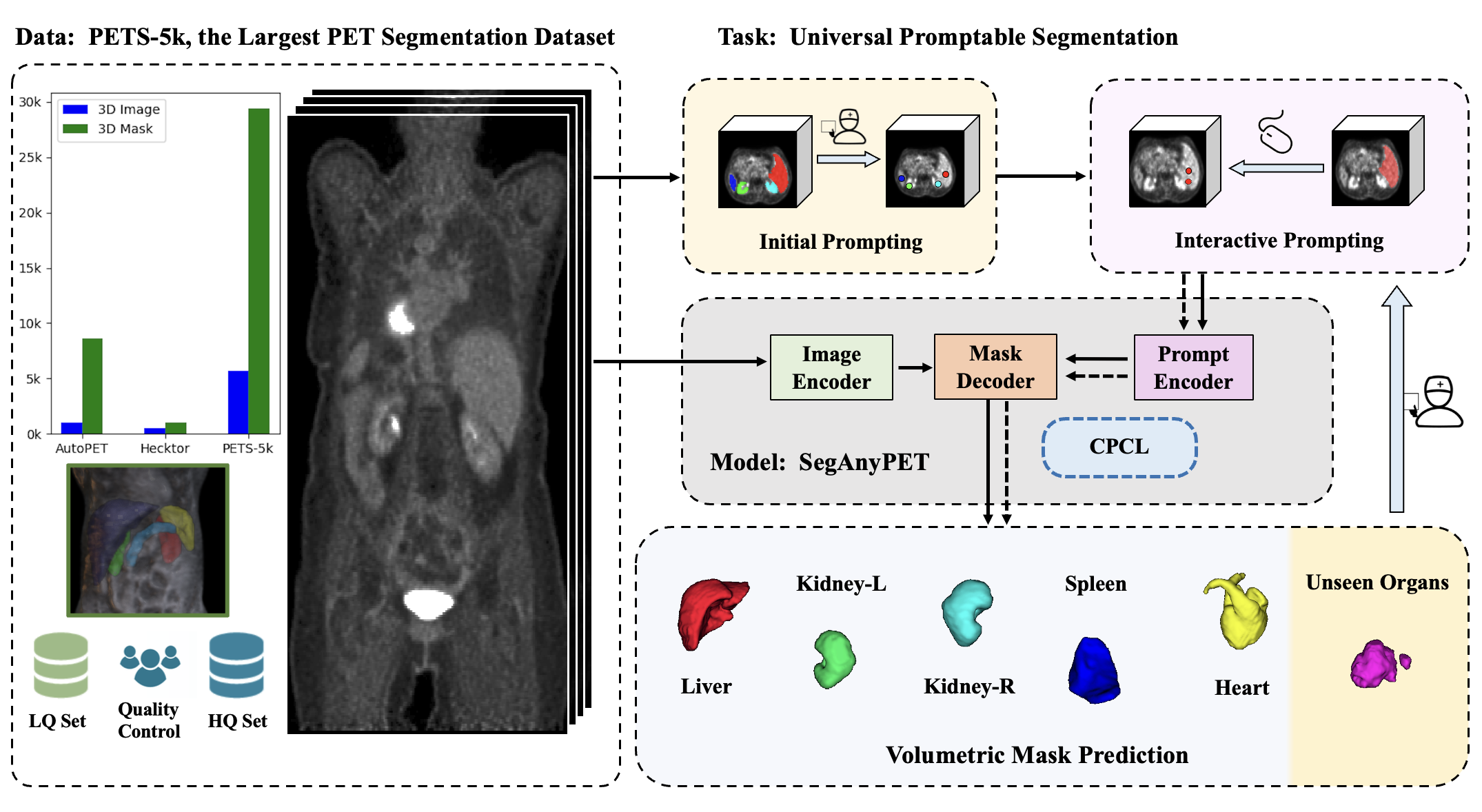}
	\caption{An overview of our work for PET segmentation foundation model. Firstly, we collect and construct PETS-5k, a large-scale PET segmentation dataset for developing the foundation model. Based on the principle of promptable segmentation, we develop SegAnyPET, a modality-specific 3D segmentation foundation model that can be efficiently adapted for universal segmentation of any target organs or lesions based on positional prompting from 3D PET images.}
	\label{Overview}
\end{figure*}

With the unprecedented developments in deep learning, deep neural networks have been widely applied and achieved great success in medical image segmentation \cite{ronneberger2015u,isensee2020nnunet}.
However, existing deep models are often tailored for specific tasks, which require task-specific training data for model development and lack the capacity for generalization across different tasks. 
The emergence of pre-trained foundation models has sparked a new era due to their remarkable generalization abilities across a wide range of downstream tasks \cite{moor2023foundation,willemink2022toward}.
For image segmentation, the introduction of the Segment Anything Model (SAM) \cite{SAM} has gained massive attention as a promptable foundation model capable of generating fine-grade segmentation masks to unseen targets using positional prompts like points or bounding boxes.
However, the original SAM is trained on natural images characterized by strong edge information, which differs significantly from medical images, especially PET images that exhibit low contrast and weak boundaries. When directly applying SAM without any adaptation, its performance significantly left behind task-specific segmentation models \cite{SAM4MIS,SAM-Empirical}.
Despite the attempts of fine-tuning and adapting SAM for medical images \cite{MedSAM,3DSAM-adapter}, these works mostly focus on CT and MRI images with detailed physiological structural information, ignoring the PET imaging. 
A significant obstacle lies in the scarcity of large-scale annotated PET datasets, which is attributable to the high costs associated with acquisition and annotation.
%Specifically, the cost of acquiring PET images is considerably higher than that of CT images.
Existing PET datasets \cite{AutoPET,hecktor} are relatively small and only focus on the segmentation of limited targets. 
Besides, the low image quality and the presence of partial volume effects complicate the annotation of segmentation targets from PET images, potentially leading to inconsistent annotation quality.

To overcome this challenge, we collect and construct a large-scale PET segmentation dataset named \textbf{PETS-5k}. The dataset consists of 5,731 three-dimensional whole-body PET images, encompassing over 1.3M 2D images, which is the largest PET dataset to date. Based on the empirical analysis, we found that existing foundation models show poor generalization performance on PET images.
To enable universal promptable segmentation from PET images, we propose \textbf{SegAnyPET}, a modality-specific foundation model that can be efficiently and robustly adapted to segment anything from PET images.
Other than the original 2D architecture of SAM, we reformulate a 3D architecture to fully utilize the inter-slice context information of 3D PET images.
To issue the challenge of discrepant annotation quality, we adopt a cross prompting confident learning (CPCL) strategy with an uncertainty-guided self-rectification process to efficiently learn from dataset with varying annotation quality for model training. The strategy does not require network modification and can be easily adapted to other promptable foundation models.
Extensive experiments demonstrate that SegAnyPET can correctly segment both seen and unseen targets using only one or few prompt points and outperform state-of-the-art foundation models and task-specific fully supervised models by a large margin, underpinning its general-purpose segmentation ability.
Our contribution can be summarized as follows:

\begin{itemize}
\item The up-to-date largest 3D PET segmentation dataset \textbf{PETS-5k}, with 5,731 three-dimensional PET images, encompassing over 1.3M 2D images, that are significantly larger than existing datasets.
\item A modality-specific foundational model \textbf{SegAnyPET} for universal promptable segmentation from PET images, demonstrating superior performance compared with state-of-the-art methods and strong generalization ability.
\item A cross prompting confident learning (CPCL) strategy for noisy-robust learning to efficiently learn from high-quality labeled data and low-quality noisy labeled data for model training.
\end{itemize}
\section{Related Work}
\label{sec:related}

\textbf{Vision Foundation Models.}
Foundation models represent a rapidly expanding field in artificial intelligence research, focusing on developing large-scale, general-purpose models with capabilities applicable across various domains and applications. Pioneering vision foundation models have been primarily based on pre-training methods such as CLIP \cite{CLIP} and ALIGN \cite{ALIGN} leverage contrastive learning techniques to train both text and image encoders. However, these models primarily excel in tasks that involve mapping images to text, such as classification.
For foundation model of image segmentation, the Segment Anything Model (SAM) \cite{SAM} represents a new paradigm of image segmentation with universal segmentation ability across different tasks and demonstrates strong potential in addressing a wide range of downstream tasks like medical image analysis \cite{zhang2024challenges,moor2023foundation}.

\textbf{Foundation Models for Medical Image Segmentation.}
As a crucial branch of image segmentation, recent studies have explored the application of SAM to the segmentation tasks from multi-modal multi-target biomedical images \cite{SAM-Empirical,SAM-SZU,ma2024segment}, while some research also focuses on fine-tuning SAM \cite{MedSAM} or adapting SAM-like foundation models \cite{SAM-Med3D,du2023segvol,wong2024scribbleprompt} on large-scale medical datasets. These studies indicate that SAM has limited generalization ability when directly applied to medical image segmentation due to the significant distribution gap, and fine-tuning SAM on medical datasets can improve the unsatisfactory performance \cite{SAM4MIS}.
In addition to the adaptation of model architectures, another line of research aims to extend the original promptable interactive segmentation method into end-to-end fully automatic segmentation by auto-prompting \cite{MedLSAM, UR-SAM} or synergy in training supervised segmentation models \cite{zhang2024semisam,li2023segment}.
However, due to the scarcity of public datasets, existing adaptions mostly focus on structural medical images like CT and MRI, ignoring molecular PET images.

\textbf{Annotation-Efficient Learning.}
Despite the outstanding performance in many medical image segmentation tasks \cite{isensee2020nnunet,wasserthal2023totalsegmentator}, the success of existing deep neural networks heavily relies on the massive radiologist-examined labeled data.
Obtaining pixel-wise annotation is time-consuming and expertise-demanding, which motivates annotation-efficient learning strategies to train deep models with limited annotations \cite{cheplygina2019not,jiao2023learning,shi2024beyond}.
Despite recent efforts of large-scale medical datasets \cite{ye2023sa,bai2024m3d}, these datasets are the integration of existing public datasets with different annotation quality. While some labels may be noisy with mistakes and could potentially hinder the learning procedure, it is important to develop noise-robust learning strategies. 
\begin{figure*}[t]
	\includegraphics[width=\linewidth]{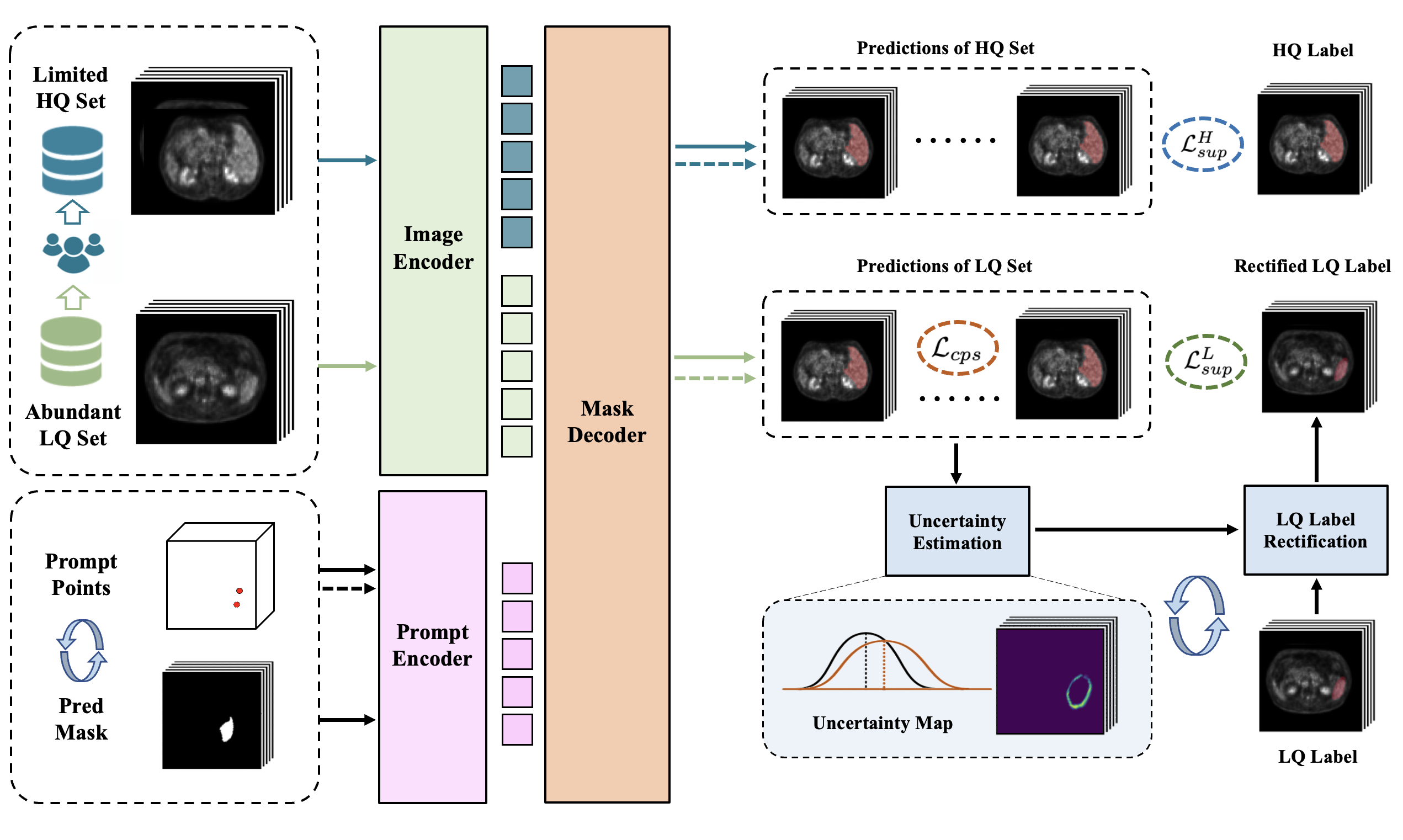}
	\caption{Illustration of the proposed cross prompting confident learning (CPCL) strategy for developing promptable segmentation foundation model based on both high quality and low quality annotations.}
	\label{CPCL}
\end{figure*}

\section{Methodology}

In this paper, we aim to develop a modality-specific foundation model for universal segmentation from PET images. An overview of our work is shown in \cref{Overview}.
Building upon the proposed large-scale PETS-5k dataset, we design a 3D promptable segmentation framework and introduce a cross prompting confident learning strategy to facilitate model training with different annotation quality.
The details are elaborated in the following sections.

\begin{algorithm}[t]
	\caption{The training procedure of SegAnyPET.}
	\label{Algorithm}
    \renewcommand{\algorithmicrequire}{\textbf{Input:}}
    \renewcommand{\algorithmicensure}{\textbf{Output:}}
	\begin{algorithmic}[1]
 		\REQUIRE{Training data $\mathcal{D}_{H}=\{X_{(i)},Y_{H(i)} \}_{i=1}^{M}$ with high-quality annotations, and $\mathcal{D}_{L}=\{X_{(i)},Y_{L(i)} \}_{i=1}^{N}$ with low-quality annotations}
		\ENSURE{SegAnyPET model parameters: Image Encoder $\mathcal{F}_{IE}$, Prompt Encoder $\mathcal{F}_{PE}$, Mask Decoder $\mathcal{F}_{D}$}
		\STATE{Initialize training details}
        \FOR{$t \Leftarrow 1$ \textbf{to} $t_{max}$}
            \STATE{Extract image embeddings $f_{img} = \mathcal{F}_{IE}(X)$}
            \STATE{Initialize $\hat Y_{0}$ as all zero matrix $\mathbf{O}$.}
            \FOR{$i \Leftarrow 1$ \textbf{to} $n_{pt}$}
                \STATE{Input prompt $p_{i}$ $\Leftarrow$ PromptGenerate($Y$, $\hat Y_{i-1}$)}
                \STATE{Update mask $\hat{Y}_{i} = \mathcal{F}_{D}(f_{img}, \mathcal{F}_{PE}(p_{i},\hat Y_{i-1}))$}
            \ENDFOR
        % \STATE{\color{teal}{\textbf{\# Cross Prompting Confident Learning.}}}
        \STATE{Average prediction $\bar Y = \frac{1}{n_{pt}}\sum\limits_{i = 1}^{n_{pt}} \hat Y_{i} $ }
        \STATE{Estimate uncertainty $U = - \sum p(\bar Y|X) \log p(\bar Y|X) $ }
        \IF{X in $\mathcal{D}_{H}$}
            \STATE{Supervised loss $\mathcal{L}^{H}_{seg} \Leftarrow (\hat{Y}_{i},Y)$ }
        \ELSIF{X in $\mathcal{D}_{L}$}
            \STATE{Label rectification $\tilde Y \Leftarrow (Y,U)$ }
            \STATE{Supervised loss $\mathcal{L}^{L}_{seg} \Leftarrow (\hat{Y}_{i},Y,Y_{rec})$ }
            \STATE{Regularization loss $\mathcal{L}_{cps} \Leftarrow (\bar Y, \{ \hat Y_{1},...,\hat Y_{n_{pt}} \})$ }
        \ENDIF
        \STATE{Total loss $\mathcal{L} = \mathcal{L}^{H}_{seg} + \lambda (\mathcal{L}_{cps} + \beta \mathcal{L}^{L}_{seg})$ }
        \STATE{Update model parameters ($\mathcal{F}_{IE},\mathcal{F}_{PE},\mathcal{F}_{D}) \Leftarrow \mathcal{L}$}
        \ENDFOR
		\RETURN model
	\end{algorithmic}
\end{algorithm}

\subsection{Network Architecture}

Motivated by the recent advance of segmentation foundation models for natural images \cite{SAM}, we reformulate a holistic 3D structure in SegAnyPET to capture the spatial information directly from volumetric images for universal segmentation.
Specifically, SegAnyPET consists of three main components, including an image encoder $\mathcal{F}_{IE}$, a prompt encoder $\mathcal{F}_{PE}$, and a mask decoder $\mathcal{F}_{D}$.
The image encoder aims to transform input images into discrete embeddings, and the prompt encoder converts input prompts into compact embeddings by combining fixed positional encoding and adaptable prompt-specific embeddings. After that, the mask decoder receives the extracted information from both the image encoder and the prompt encoder and incorporates prompt self-attention and cross-attention in two directions for prompt-to-image and image-to-prompt attention to update the image embeddings and prompt embeddings. The processed feature map is then up-sampled and passes through a multi-layer perception to generate the output segmentation masks. 
Details of the network components are shown in the Appendix.
Given an input image $X$ and spatial prompt $p$, the output segmentation mask $\hat Y$ can be formally expressed as:

\begin{equation}
\hat Y_{i} = \left\{
\begin{array}{lllll}
\mathbf{O}_{H \times W \times D}, & i=0 \\ \\
\mathcal{F}_{D}(\mathcal{F}_{IE}(X), \mathcal{F}_{PE}(p_{i},\hat Y_{i-1})), & i \ge 1 \\
\end{array}\right.
\end{equation} 
where $\mathbf{O}_{H \times W \times D}$ is an all-zero matrix as the pre-defined initial segmentation mask. In the training loop, new point prompts are generated from error regions based on the previous segmentation mask to simulate the manual interactive segmentation workflow.

\begin{table*}[t]
	\centering
    \normalsize
    \setlength\tabcolsep{10pt}
	\renewcommand\arraystretch{1.15}
	\begin{tabular}{c|c|ccccc|c}
		\hline 	\hline
		\multirow{2}{*}{Method}  &  \multirow{2}{*}{Prompt} & \multicolumn{6}{c}{Organ Segmentation DSC performance [\%]}  \\
\cline{3-8}  &&  Liver & Kidney-L & Kidney-R & Heart & Spleen & Avg  \\ \hline
SAM  \cite{SAM} & N points & 26.55 & 9.38 & 9.10 & 14.44 & 6.30 & 13.15 \\
MedSAM  \cite{MedSAM} & N points & 0.25 & 0.19 & 1.32 & 0.27 & 0.27 & 0.46 \\
SAM-Med3D \cite{SAM-Med3D} & 1 point & 51.63 & 21.01 & 19.17 & 60.11 & 25.41 & 35.46 \\
SAM-Med3D-organ & 1 point & 80.25 & 44.70 & 35.76 & 74.00 & 69.23 & 60.79 \\ 
SAM-Med3D-turbo & 1 point & 79.46 & 66.95 & 72.81 & 73.03 & 68.19 & 72.09 \\ \hline
\rowcolor{gray!25} \textbf{SegAnyPET}  & 1 point & 93.06 & 89.84 & 90.61 & 88.29 & 90.67 & 90.49 \\ \hline
SAM  \cite{SAM} & 3N points & 43.85 & 23.21 & 22.16 & 29.09 & 11.83 & 26.03  \\
MedSAM  \cite{MedSAM} & 3N points & 26.59 & 28.86 & 28.98 & 18.82 & 32.96 & 27.24 \\
SAM-Med3D \cite{SAM-Med3D} & 3 points & 62.15 & 28.21 & 31.19 & 61.44 & 27.07 & 42.01 \\
SAM-Med3D-organ & 3 points & 84.82 & 47.33 & 48.57 & 75.85 & 74.60 & 66.23 \\
SAM-Med3D-turbo & 3 points & 84.11 & 74.05 & 76.17 & 75.24 & 73.34 & 76.58 \\ \hline
\rowcolor{gray!25} \textbf{SegAnyPET}  & 3 points & 93.36 & 90.25 & 90.95 & 88.86 & 91.10 & 90.90 \\ \hline
SAM  \cite{SAM} & 5N points & 54.49 & 47.16 & 37.42 & 42.19 & 18.79 & 40.01 \\
MedSAM  \cite{MedSAM} & 5N points & 36.53 & 37.53 & 39.22 & 24.71 & 41.30 & 35.86 \\
SAM-Med3D \cite{SAM-Med3D} & 5 points & 61.05 & 31.05 & 31.98 & 61.88 & 29.75 & 43.14 \\
SAM-Med3D-organ & 5 points & 85.52 & 49.56 & 54.40 & 76.30 & 75.13 & 68.18 \\
SAM-Med3D-turbo & 5 points & 85.56 & 76.74 & 78.08 & 76.16 & 75.20 & 78.35 \\ \hline
\rowcolor{gray!25} \textbf{SegAnyPET}  & 5 points & 93.42 & 90.39 & 91.24 & 88.95 & 91.22 & 91.05 \\ \hline \hline
SegResNet \cite{segresnet} & Auto & 92.28 & 85.56 & 82.94 & 88.21 & 86.55 & 87.11 \\
SwinUNETR \cite{hatamizadeh2021swin} & Auto & 93.03 & 89.23 & 83.62 & 89.03 & 87.93 & 88.57 \\
nnUNet \cite{isensee2020nnunet}  & Auto & 93.16 & 90.12 & 86.60 & 90.96 & 88.32 & 89.83 \\
STUNet \cite{huang2023stunet} & Auto & 93.15 & 90.11 & 85.60 & 90.38 & 88.83 & 89.61 \\
\hline  \hline
	\end{tabular}
 	\caption{Comparison of DSC performance [\%] with state-of-the-art segmentation foundation models for \textbf{\underline{zero-shot interactive}} segmentation  and task-specific models for \textbf{\underline{training-based automatic}} segmentation from PET images. N denotes the count of slices containing the target object (N ranges from 20 to 50 in our task).} \label{Table_Seen}
\end{table*}

\subsection{Cross Prompting Confident Learning}

Due to the dependency of domain knowledge for annotating medical images, acquiring large amounts of high-quality labeled data is infeasible.
A prevalent approach to mitigating this issue is collecting additional labeled data with varying annotation qualities like crowdsourcing from non-experts or using model-generated labels.
Given that some annotations may be fraught with noise or inaccuracies, which could impede the learning process, it is imperative to devise noise-robust learning strategies to efficiently learn from both expert-examined high-quality annotations (termed as HQ Set) and low-quality annotations with possible mistakes (termed as LQ Set).
To issue the challenge of discrepant annotation quality, we adopt a cross prompting confident learning (CPCL) strategy with a cross prompting consistency similarity regularization and an uncertainty-guided self-rectification process.
\Cref{CPCL} depicts the workflow of CPCL. The utilization of LQ set can be mainly divided into the following two parts.

\textbf{Exploit Knowledge Inside the Images.}
Given that LQ labels could be the encumbrance for model training, it is natural to adopt the semi-supervised learning setting that casts LQ labels away and exploits the knowledge inside the image data alone.
For promptable segmentation settings, the output heavily relies on the input prompts, and different prompts may cause variances in the segmentation results even when they refer to the same object given the same image. 
This phenomenon underscores the imperative to bolster the invariance of segmentation outputs across different prompts, which could be utilized as unsupervised regularization for training.
Based on this motivation, we harness the capabilities inherent to the prompting mechanism and develop a cross prompting consistency strategy to learn from unlabeled data.
Within each training loop, we estimate the unsupervised regularization loss as follows:

\begin{equation}
\bar Y = \frac{1}{n_{pt}}\sum\limits_{i = 1}^{n_{pt}} \hat Y_{i} 
\end{equation}
\begin{equation}
\mathcal{L}_{cps} = \sum_{i=1}^{n_{pt}} \| \bar Y - \hat Y_{i} \| ^{2}
\label{cps}
\end{equation}
where $n_{pt}$ represents the setting of prompt point number in each training loop and $\bar Y$ represents the averaged segmentation result in the loop.

\textbf{Self-Rectification Towards Effective Guidance.}
In addition to image-based information, effectively leveraging concomitant noisy labels inherent in LQ set is crucial to further improve the performance. To this end, we further propose an uncertainty-based self-rectification process to alleviate the potential misleading caused by label noises.
Uncertainty estimation refers to the process of predicting the uncertainty or confidence of model predictions by assigning a probability value that represents the self-confidence associated with the prediction, so as to quantify the reliability of the output and to identify when the network may not be performing well.
For image segmentation, the model assigns a probability value that represents the self-confidence associated with each pixel's prediction, and lower confidence (\textit{i.e.} higher uncertainty) can be a heuristic likelihood of being mislabeled pixels.

Building upon the foundation of promptable segmentation, instead of adding perturbations to input images or models, we leverage the advantage of the prompting mechanism and approximate the segmentation uncertainty with different input prompts in the training loop. Following \cite{zhang2023uncertainty}, we select the predictive entropy to approximate the uncertainty since it has a fixed range. The aleatoric uncertainty of a given image is formulated as follows:

\begin{equation}
U = - \sum p(\bar Y|X) \log p(\bar Y|X)
\label{unc}
\end{equation}

To exploit informative knowledge from LQ labels, we perform an uncertainty-based self-rectification process for noisy label refinement.
With the guidance of estimated uncertainty, we filter out relatively unreliable regions with higher uncertainty as potential error maps. Then the rectification operation is formulated as:
\begin{equation}
\tilde y_{i} = y_{i} + \mathbb I(u_{i}>H) \cdot (-1)^{y_{i}}
\label{rec}
\end{equation}
where $\mathbb I(.)$ is the indicator function, $y_{i}$ and $u_{i}$ are the initial annotation and estimated uncertainty at i-th voxel. Then the rectified label of the whole volume $\tilde Y$ is $\{\tilde y\} \in \{0,1\}_{H \times W \times D}$.
For LQ set, we utilize rectified label for supervised training instead of the original noisy label.
With the help of the label rectification process, the negative effects brought by label noises can be eliminated to some extent and provide more rewarding guidance.

\subsection{Overall Training Procedure}

\Cref{Algorithm} presents the overall training procedure of SegAnyPET. The total loss for model training is weighted combination of the supervised loss $\mathcal{L}^{H}_{seg}$ on HQ set, the unsupervised regularization loss $\mathcal{L}_{cps}$ on LQ set, and the rectified supervised loss $\mathcal{L}^{L}_{seg}$ on LQ set, calculated by:

\begin{equation}
\mathcal{L} = \mathcal{L}^{H}_{seg} + \lambda (\mathcal{L}_{cps} + \beta \mathcal{L}^{L}_{seg})
\end{equation}
where $\beta$ is a fixed weighting coefficient, and $\lambda$ is a ramp-up trade-off weighting coefficient to avoid domination by misleading targets at the early training stage.

\begin{table*}[t]
	\centering
    \normalsize
    \setlength\tabcolsep{5pt}
	\renewcommand\arraystretch{1.15}
	\begin{tabular}{c|c|ccccccc|c}
		\hline 	\hline
		\multirow{2}{*}{Method}  &  \multirow{2}{*}{Prompt} & \multicolumn{8}{c}{Training Invisible Organ Segmentation DSC Performance [\%]}  \\
\cline{3-10}  && Aorta & Lung-LL & Lung-LR & Lung-UL & Lung-UR & Lung-MR & Prostate & Avg  \\ \hline
SAM  \cite{SAM} & N points & 2.55 & 9.60 & 12.24 & 11.17 & 15.23 & 9.75 & 8.01 & 9.79 \\
MedSAM  \cite{MedSAM} & N points & 0.10 & 0.89 & 0.70 & 0.97 & 1.70 & 2.57 & 0.51 & 1.06  \\
SAM-Med3D \cite{SAM-Med3D} & 1 point & 11.29 & 25.60 & 45.95 & 37.94 & 47.15 & 31.64 & 9.93 & 29.93 \\
SAM-Med3D-organ & 1 point & 24.12 & 45.71 & 42.16 & 56.32 & 61.71 & 37.85 & 32.25 & 42.87  \\
SAM-Med3D-turbo & 1 point & 21.28 & 31.55 & 37.82 & 44.36 & 53.56 & 39.34 & 25.46 & 36.20 \\ \hline
\rowcolor{gray!25} \textbf{SegAnyPET}  & 1 point & 82.99 & 88.32 & 89.76 & 89.47 & 91.91 & 89.17 & 91.67 & 89.04 \\ \hline
SAM  \cite{SAM} & 3N points & 6.45 & 13.91 & 16.88 & 16.40 & 22.04 & 15.45 & 19.30 & 15.78 \\
MedSAM  \cite{MedSAM} & 3N points & 17.95 & 24.85 & 20.18 & 18.29 & 21.23 & 23.86 & 25.82 & 21.74 \\
SAM-Med3D \cite{SAM-Med3D} & 3 points & 13.38 & 29.77 & 46.62 & 45.46 & 50.50 & 36.93 & 13.73 & 33.77 \\
SAM-Med3D-organ & 3 points & 28.76 & 48.77 & 50.06 & 64.17 & 69.10 & 51.05 & 33.84 & 49.39 \\
SAM-Med3D-turbo & 3 points & 27.37 & 36.55 & 47.00 & 57.91 & 66.17 & 57.66 & 31.21 & 46.27 \\ \hline
\rowcolor{gray!25} \textbf{SegAnyPET}  & 3 points & 84.51 & 89.09 & 90.54 & 90.26 & 92.52 & 89.72 & 92.19 & 89.83 \\ \hline
SAM  \cite{SAM} & 5N points & 14.58 & 19.37 & 23.37 & 22.97 & 31.62 & 24.88 & 41.03 & 25.40 \\
MedSAM  \cite{MedSAM} & 5N points & 21.73 & 27.45 & 25.55 & 23.05 & 27.91 & 28.24 & 37.69 & 27.37 \\
SAM-Med3D \cite{SAM-Med3D} & 5 points & 14.07 & 37.54 & 48.46 & 50.68 & 51.01 & 38.43 & 13.37 & 36.23 \\
SAM-Med3D-organ & 5 points & 29.95 & 53.12 & 54.02 & 65.87 & 71.35 & 52.55 & 35.25 & 51.73 \\
SAM-Med3D-turbo & 5 points & 31.92 & 42.05 & 48.36 & 64.34 & 69.65 & 61.15 & 32.77 & 50.03 \\ \hline
\rowcolor{gray!25} \textbf{SegAnyPET}  & 5 points & 84.63 & 89.21 & 90.69 & 90.34 & 92.63 & 89.80 & 92.57 & 89.98 \\ \hline \hline
	\end{tabular}
 	\caption{Generalization performance to \textbf{\underline{training invisible organs}} with comparison to state-of-the-art segmentation foundation models for zero-shot interactive segmentation from PET images.} \label{Table_Unseen}
\end{table*}

\section{Experiments}

\subsection{Datasets and Experimental Setup}

We conduct comprehensive experiments on two different PET datasets.
The first dataset is the proposed \textbf{PETS-5k dataset}, which consists of 5,731 three-dimensional whole-body PET images with the annotation of 5 target organs.
% including liver, left kidney, right kidney, heart, and spleen. 
Among the dataset, a subset of 100 images are with high-quality annotations verified by domain experts, which is split into 40 cases serving as HQ training set and 60 cases serving as internal test set.
The remaining 5,631 images are noisy-annotated with possible mislabeled or unlabeled pixels serving as the LQ training set.
To evaluate the generalization performance for universal segmentation, we annotate 7 additional training invisible organs from the internal test set.
% including aorta, prostate, left lung lower lobe, right lung lower lobe, left lung upper lobe, right lung upper lobe, and right lung middle lobe from the internal test set.
The second dataset is \textbf{AutoPET-Organ dataset} extended from the AutoPET dataset  \cite{AutoPET} consists of 1,014 three-dimensional whole-body PET images with corresponding annotation of tumor lesions. To evaluate the generalization ability, we further annotate a small subset with 100 cases with all 12 organs as an external test set.
Following the setting for promptable segmentation, the input prompts are simulated based on the ground truth mask with random perturbations. We compare proposed SegAnyPET with state-of-the-art segmentation foundation models \cite{SAM,MedSAM,SAM-Med3D} and training-based task-specific models \cite{segresnet,isensee2020nnunet,hatamizadeh2021swin,huang2023stunet}. 
More information of the datasets and implementation details are shown in the Appendix.

\subsection{Experiments}

\begin{figure}
    \centering
	\includegraphics[width=\linewidth]{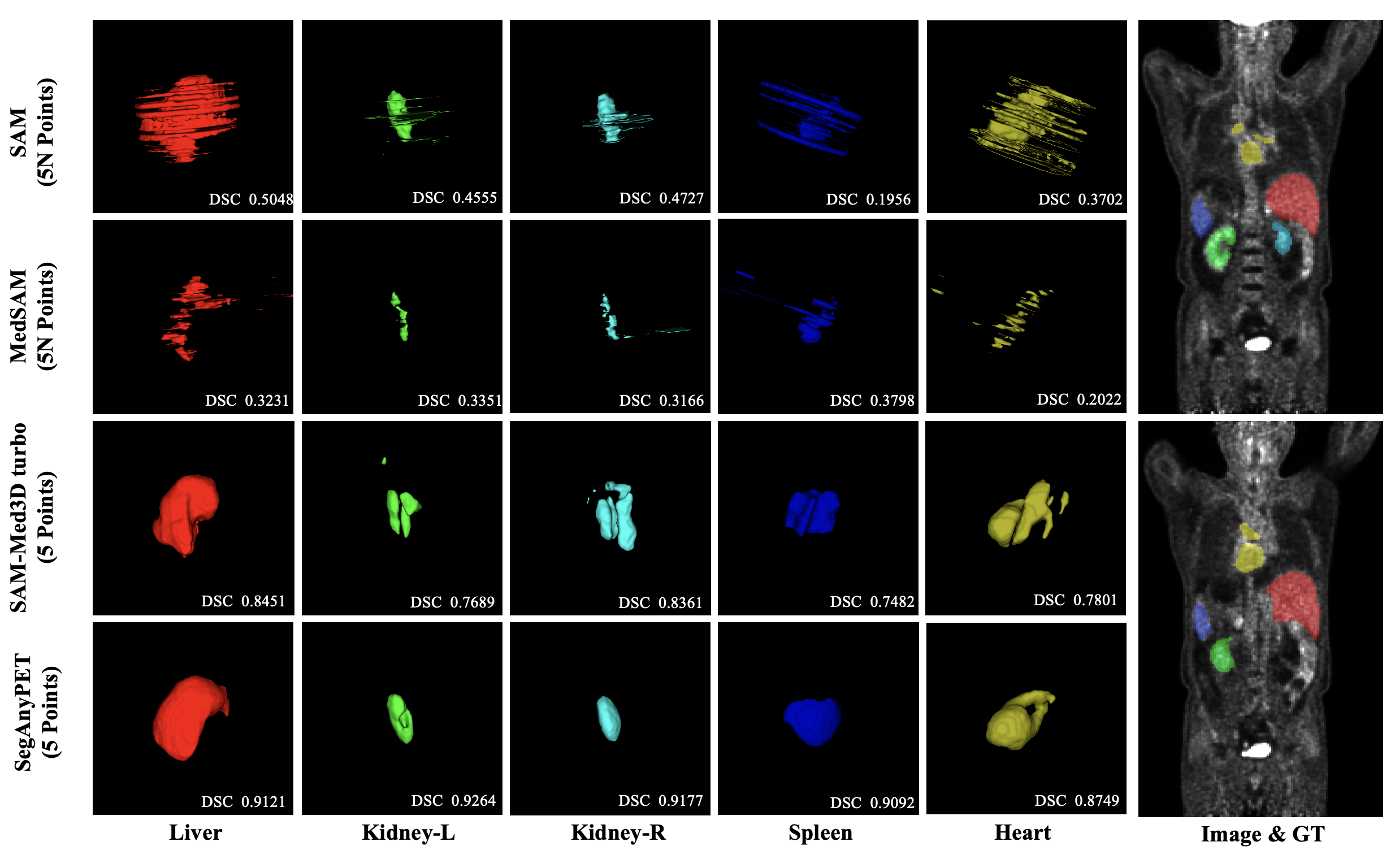}
	\caption{Visual comparison of segmentation results of different promptable segmentation foundation models for PET organ segmentation.}
    \label{organseg}
\end{figure}

\textbf{Comparison with Universal Foundation Models.}
The original SAM \cite{SAM} and MedSAM \cite{MedSAM} are designed for 2D segmentation tasks and cannot handle 3D inputs directly,
necessitating input prompts for each 2D slice containing the target to conduct slice-by-slice segmentation.
In contrast, SegAnyPET and other 3D SAM medical adaptions \cite{SAM-Med3D} can be directly utilized to segment the target organs from input volume with one or few prompts. 
\Cref{Table_Seen} presents the performance of SegAnyPET with state-of-the-art foundation models on our internal test set under different prompt settings.
We can observe that 2D models exhibit lower performance due to the lack of volumetric spatial information. In contrast, 3D models can achieve better performance with less manual prompting effort. Compared with state-of-the-art methods, SegAnyPET exhibits significantly better segmentation performance and outperforms the second-ranked method by a significant improvement up to 18.40\% on average DSC. 
We visualize the segmentation results of SegAnyPET and other segmentation foundation models for organ segmentation in \cref{organseg}, which intuitively show that SegAnyPET performs more precise segmentation with higher DSC.

\textbf{Comparison with Task-Specific Models.}
In addition to foundation models, we also conduct a comparison with state-of-the-art task-specific models in \cref{Table_Seen}. 
These task-specific models are trained on the HQ training set focusing on the segmentation task of target organs. When adapting to training invisible organs, additional annotation and model training are required for segmentation.
Despite the manual prompting efforts, we observe that SegAnyPET significantly outperforms task-specific models, while preserving the generalization ability for universal segmentation.

\begin{table*}[h]
	\centering
    \small
    \setlength\tabcolsep{3pt}
	\renewcommand\arraystretch{1.3}
	\begin{tabular}{c|ccc|ccc|ccc|ccc}
		\hline 	\hline
Method & \multicolumn{3}{c}{SAM \cite{SAM}} & \multicolumn{3}{|c}{MedSAM \cite{MedSAM}} &  \multicolumn{3}{|c}{SAM-Med3D \cite{SAM-Med3D}}  &  \multicolumn{3}{|c}{\textbf{SegAnyPET}}   \\ \hline
Prompt & N points & 3N points & 5N points & N points & 3N points & 5N points &  1 point & 3 points & 5 points & 1 point & 3 points & 5 points \\
 \hline
Liver & 31.02 & 38.24 & 51.51 & 3.91 & 37.18 & 48.19 & 59.06 & 73.44 & 78.47 & 76.70  & 83.01 & 83.75\\ 
Kidney-L & 6.89 & 9.80 & 18.65 & 0.63 & 22.53 & 26.78 & 67.64 & 71.93 & 70.86 & 75.97  & 77.36 & 77.86 \\ 
Kidney-R & 7.08 & 9.97 & 17.69 & 1.08 & 25.82 & 33.37 & 54.82 & 62.57 & 63.71 & 71.56  & 73.95 & 75.25\\ 
Heart & 18.79 & 23.06 & 30.93 & 0.78 & 29.43 & 32.15 & 48.91 & 53.84 & 55.14 & 67.62 & 70.95 & 71.64 \\ 
Spleen & 11.05 & 15.14 & 23.94 & 0.74 & 30.52 & 32.53 & 37.58 & 43.59 & 49.69 & 77.97 & 80.16 & 80.84 \\ \hline
\rowcolor{gray!15} Aorta & 2.81 & 4.00 & 7.69 & 1.53 & 23.34 & 24.24 & 19.79 & 24.07 & 27.47 & 16.00 & 18.73 & 22.57 \\ 
\rowcolor{gray!15} Lung-LL & 13.16 & 15.49 & 21.93 & 2.84 & 21.81 & 22.19 & 32.27 & 38.05 & 41.77 & 13.32 & 24.09 & 26.73\\ 
\rowcolor{gray!15} Lung-LR & 16.49 & 19.45 & 26.11 & 1.65 & 26.48 & 28.52 & 45.18 & 47.99 & 49.08 & 26.67 & 37.87 & 41.35 \\ 
\rowcolor{gray!15} Lung-UL & 15.18 & 18.38 & 26.42 & 1.48 & 22.18 & 23.33 & 51.69 & 60.23 & 64.18 & 10.80 & 18.04 & 19.14 \\ 
\rowcolor{gray!15} Lung-UR & 18.36 & 21.65 & 29.13 & 1.70 & 29.10 & 33.74 & 41.31 & 48.86 & 49.92 & 19.08 & 39.95 & 43.34 \\ 
\rowcolor{gray!15} Lung-MR & 11.94 & 15.32 & 21.11 & 3.26 & 29.52 & 30.25 & 28.55 & 37.04 & 42.76 & 16.36 & 25.72 & 28.69 \\
\rowcolor{gray!15} Prostate & 3.96 & 6.60 & 17.51 & 0.96 & 23.80 & 29.71 & 31.52 & 43.11 & 43.48 & 35.93 & 38.47 & 39.87 \\ 
\hline \hline
	\end{tabular}
	\caption{Generalization performance to \textbf{\underline{unseen out-of-distribution}} AutoPET-Organ dataset with comparison to state-of-the-art segmentation foundation models for zero-shot interactive segmentation from PET images.} \label{Table_AutoPET}
\end{table*}

\textbf{Generalization to Unseen Organs and Dataset.}
To validate the generalization performance of SegAnyPET, we conduct additional experiments of foundation models for promptable segmentation of training invisible organs. As shown in \cref{Table_Unseen}, we observe SegAnyPET also exhibits strong generalization performance and outperforms state-of-the-art segmentation foundation models. The experimental results demonstrate that by efficiently learning representations from large-scale PET images, SegAnyPET exhibits comparable performance for training invisible organs. 
Furthermore, we evaluate the proposed method on training invisible AutoPET-Organ dataset with significant distribution gap compared with PETS-5k dataset. As shown in \cref{Table_AutoPET}, we observe that SegAnyPET can achieve satisfying generalization ability on unseen dataset.

\textbf{Ablation Analysis.}
In \cref{Table_Ablation}, we conduct ablation experiments to evaluate the effectiveness of proposed training strategy. 
Three different training strategies are conducted including fine-tuning with mixed HQ and LQ, training with additional unsupervised regularization for LQ, and the proposed cross prompting confident learning strategy. Experimental results show that the proposed strategy can achieve better performance on both seen and unseen targets.

\section{Discussion and Conclusion}
In this work, we propose SegAnyPET, a modality-specific foundation model for universal promptable segmentation from 3D PET images.
To this end, we collect and construct a large-scale PET segmentation dataset, PETS-5k, which consists of 5,731 three-dimensional whole-body PET images, encompassing over 1.3M 2D images to train the foundation model.
To facilitate the model training with different annotation quality, we design a cross prompting confident learning strategy building on the foundation of promptable segmentation for noise-robust learning.
Through comprehensive evaluations on both internal and external datasets,
our model shows substantial capabilities in segmenting both seen and unseen targets using only one or few prompt points and robust generalization abilities to manage new data and tasks.
Its performance significantly exceeds both state-of-the-art segmentation foundation models and task-specific fully supervised models.

\begin{table}[h]
    \centering
    \normalsize
    \setlength\tabcolsep{9pt}
	\renewcommand\arraystretch{1.1}
	\begin{tabular}{c|c|c|c}
		\hline 	\hline
		Strategy  & Prompts & Seen & Unseen \\ \hline
\multirow{3}{*}{Fine-Tuning}  & 1 point & 87.61 & 62.74  \\
& 3 points & 88.02 & 63.65 \\
& 5 points & 88.13 & 63.94 \\ \hline 
\multirow{3}{*}{Consistency} & 1 point & 89.28 & 67.52 \\
& 3 points & 89.46 & 70.46 \\
& 5 points & 89.49 & 70.97 \\ \hline 
\multirow{3}{*}{CPCL} & 1 point & 90.49 & 73.96   \\ 
& 3 points & 90.90 & 77.09 \\
& 5 points & 91.05 & 77.87 \\  \hline \hline
	\end{tabular}
	\caption{Ablation analysis of different training strategies for zero-shot interactive segmentation from PET images. } \label{Table_Ablation}
\end{table}

While SegAnyPET boasts strong capabilities, there are still inherent limitations due to manual efforts for promptable segmentation, which could potentially reduce the usability of the model compared with fully automatic segmentation. 
We aim to incorporate semantic information together with positional prompts as additional support to enable efficient and precise auto-prompting for automatic segmentation in the future.
Besides, despite the differences between existing structural medical images, these large-scale medical datasets \cite{bai2024m3d} can still provide useful guidance for training in conjunction with PET images.
In conclusion, as the first foundation model for PET images, we believe that SegAnyPET will advance the developments of universal segmentation and serve as a powerful pre-trained model to facilitate the applications to various downstream tasks for molecular imaging.

{
    \small
    \bibliographystyle{ieeenat_fullname}
    \bibliography{main}
}

\clearpage
\setcounter{page}{1}
\maketitlesupplementary

\begin{table*}[t]
	\centering
    \normalsize
    \setlength\tabcolsep{7pt}
	\renewcommand\arraystretch{1.3}
	\begin{tabular}{c|ccc|cc}
		\hline
Dataset & Split   & Annotation Targets & Scans & New Data & New Label \\ \hline
 & Train LQ &  5 target organs & 5,631 & $\checkmark$ & $\checkmark$ \\ 
PETS-5k & Train HQ &  5 target organs & 40 & $\checkmark$ & $\checkmark$\\ 
 & Internal Test & all 12 organs & 60 & $\checkmark$ & $\checkmark$\\ \hline
AutoPET & External Test & tumor lesion & 1,014 \\ \hline
AutoPET-Organ & External Test & all 12 organs & 100 &&$\checkmark$ \\ \hline 
\end{tabular}
 \caption{Information summary of datasets involved in the construction and evaluation of SegAnyPET.} \label{Table_Dataset}
\end{table*}

\section*{A. Dataset Information}
\label{app_data}

In this work, our experiments are conducted on one private dataset (PETS-5k) and one public dataset (AutoPET) consisting of 3D whole-body $^{18}$F-fluorodeoxyglucose positron emission tomography ($^{18}$F-FDG-PET) images, which is the most widely used PET tracer in oncology.
As a non-specific tracer, $^{18}$F-FDG can be used for whole-body imaging to reflect tissue glucose metabolism, which makes the imaging useful in assessing the systemic distribution and metastasis of tumors.
Organ segmentation from $^{18}$F-FDG-PET images can be used to evaluate differences in the maximum standardized uptake values (SUVmax) of different organs, thereby assisting in the diagnosis of malignant tumors.

\subsection*{A.1. PETS-5k Dataset}

The proposed PETS-5k dataset consists of 5,731 three-dimensional whole-body $^{18}$F-FDG PET images collected from one local medical center. 
Patients were fasted for at least 6h and had a blood glucose level $<$ 200 mg/dL before the PET/CT examination. PET/CT imaging was performed at a median uptake time of 67 min (range from 53 to 81 min) after intra-venous injection of $^{18}$F-FDG (3.7 MBq/kg).
All data were acquired on PET/CT scanners (Siemens Biograph mCT) with 5 min per bed position. A low-dose CT scan (120 kVp; 40–100 mAs; 5 mm slice thickness) was performed from the upper thigh to the skull base, followed by a PET scan with 3D Flowmotion acquisition mode. PET images were reconstructed with 4.07 $\times$ 4.07 $\times$ 3 $mm^{3}$ voxels using CT-based attenuation correction by Siemens-specific TrueX algorithm.

\subsection*{A.2. AutoPET Dataset}

The public AutoPET dataset consists of 1,014 three-dimensional whole-body $^{18}$F-FDG PET images.
All data were acquired using cutting-edge PET/CT scanners, including the Siemens Biograph mCT, mCT Flow, and Biograph 64, as well as the GE Discovery 690. These scans were conducted following standardized protocols in alignment with international guidelines. The dataset encompasses whole-body examinations, typically ranging from the skull base to the mid-thigh level. More details can be found in the original paper \cite{AutoPET}.

\subsection*{A.3. Organ Selection and Annotation}

Due to the characteristics of molecular imaging, some target anatomical structures in segmentation tasks structural images like CT and MRI may not be apparent in PET images. As a result, we select out five most clinical-important target organs for training, including liver, left kidney, right kidney, heart, and spleen. 
To evaluate the model performance on training invisible organs, we further annotate seven organs in the internal test set including aorta, prostate, left lung lower lobe, right lung lower lobe, left lung upper lobe, right lung upper lobe, and right lung middle lobe.
These additional organs are not used for model training and only used as test set to evaluate of the generalization performance of SegAnyPET for universal segmentation of unseen targets.

For PETS-5k dataset, all the images are preliminary annotated by developed state-of-the-art segmentation model and one junior annotator using LIFEx v7.6.0 \cite{nioche2018lifex}. Among the dataset, 100 cases are then checked and refined by two senior experts, which serve as the HQ training set and test set, while the remaining cases serve as the LQ set in our task.
For AutoPET dataset, the original task is only focused on tumor lesion segmentation. In addition to the original tumor annotation, we select out and annotate a small subset of 100 cases to annotate all the 12 target organs, named AutoPET-Organ. The AutoPET-Organ is used as an external test set to evaluate the generalization performance of SegAnyPET on training invisible dataset.

\subsection*{A.4. Summary and Visualization}

\begin{figure*}[t]
    \centering
	\includegraphics[width=0.98\linewidth]{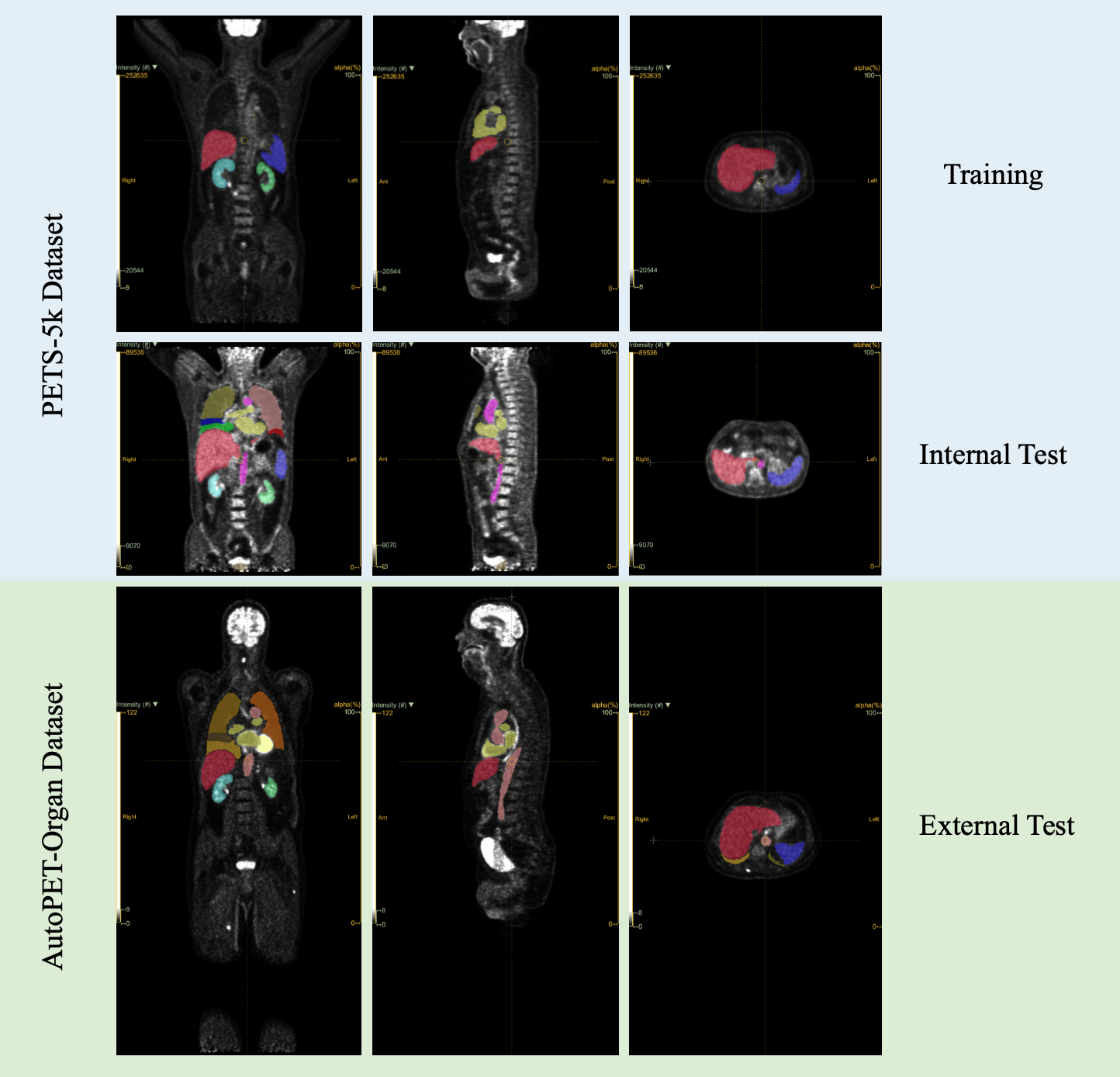}
	\caption{Visualization of PET images and corresponding organ annotations of PETS-5k dataset and AutoPET-Organ dataset.}
    \label{data_vis}
\end{figure*}

An overview information summary and visualization of the datasets used in our work are shown in \cref{Table_Dataset} and \cref{data_vis}.

\begin{figure*}[t]
    \centering
	\includegraphics[width=\linewidth]{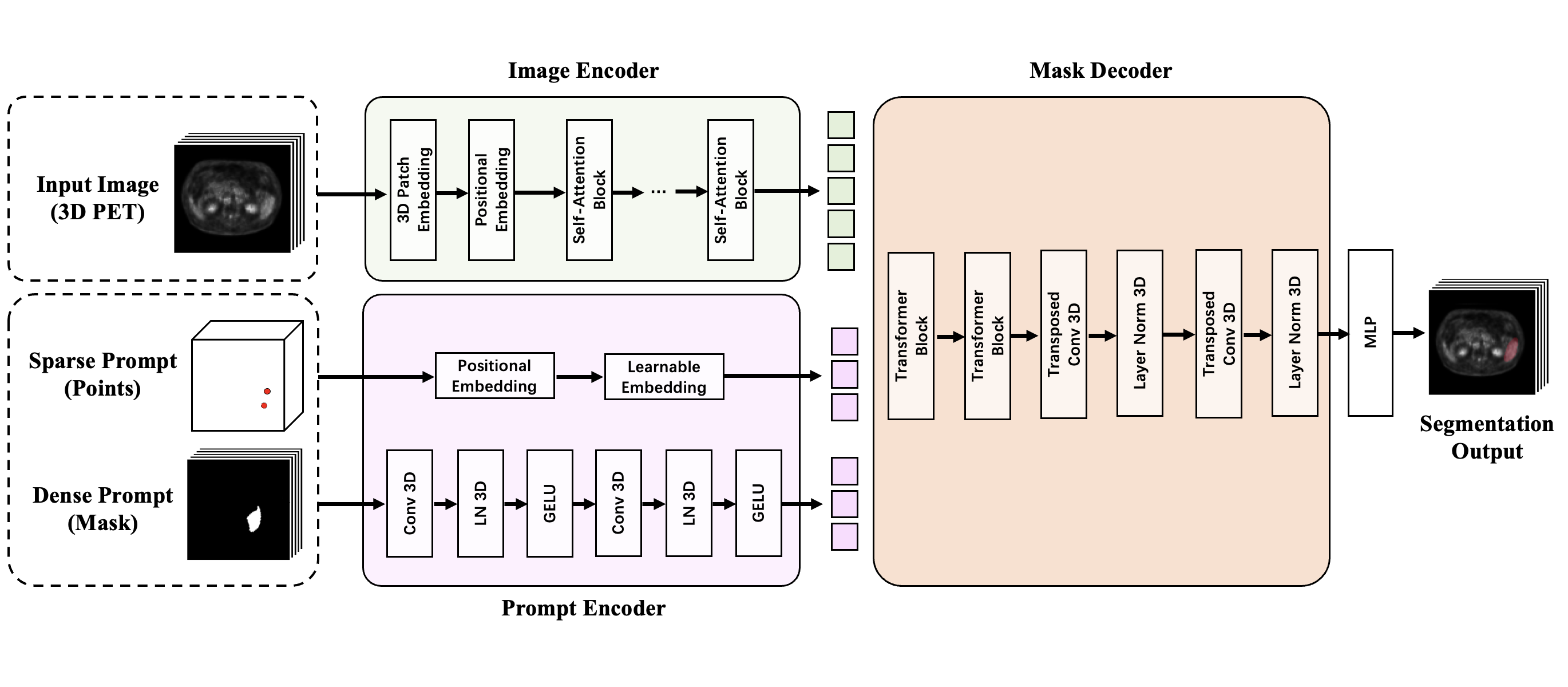}
	\caption{The detailed architecture of the network components of SegAnyPET.}
    \label{structure}
\end{figure*}

\section*{B. Methodological Details}

\subsection*{B.1. SegAnyPET Architecture}
\label{app_architecture}

The detailed architecture of SegAnyPET is shown in \cref{structure}.
Following the design in \cite{SAM-Med3D}, for the image encoder, the input patch size is set to 16$\times$16$\times$16 with a patch embedding dimension of 768, paired with a learnable 3D absolute positional encoding.
Then the embeddings of patches are input to 3D self-attention blocks. The depth of self-attention blocks is set to 16.
Within the prompt encoder, sparse prompts are leverage by 3D position embedding to represent 3D spatial differences, while dense prompts are handled with 3D convolutions followed by layer normalization and GELU activation.
The mask decoder is integrated with 3D upscaling procedures, employing 3D transformer blocks and 3D transposed convolutions to get the final segmentation result.

\subsection*{B.2. Implementation Training Details}

Our method is implemented in Python with PyTorch and trained on 4 NVIDIA Tesla A100 GPUs, each with 80GB memory. 
We use the AdamW optimizer with an initial learning rate of 0.0008 and a weight decay factor of 0.1.
The training was performed for a total of 200 epochs on the constructed PETS-5k dataset.
The batch size is set to 12 with a volumetric input patch size of 128$\times$128$\times$128. 
To handle the learning rate schedule, we employed the MultiStepLR scheduler, which adjusts the learning rate in predefined steps with 120 and 180 epochs, with a gamma value of 0.1, indicating that the learning rate is reduced by 10\% of its original value at each step.
In distributed training scenarios, we utilized gradient accumulation with 20 steps to simulate larger effective batch sizes, which can improve model performance by providing a more accurate estimate of the gradient.
% \textbf{For manual prompting interactive segmentation task, the input prompts are simulated based on the ground truth.}
For total loss in the training loop, the ramp-up trade-off weighting coefficient $\lambda$ is scheduled by the time-dependent Gaussian function as $ \lambda=\omega_{max}*e^{-5(1-t/t_{max})}$, where $t_{max}$ is the maximum training iteration, $\omega_{max}$ is the maximum weight set as 0.1 and $\beta$ is set to 5.
The weighting coefficient can avoid the domination by misleading targets at the early training stage.

\subsection*{B.3. 2D/3D Prompt Generation Strategy}

\begin{figure*}[t]
    \centering
	\includegraphics[width=\linewidth]{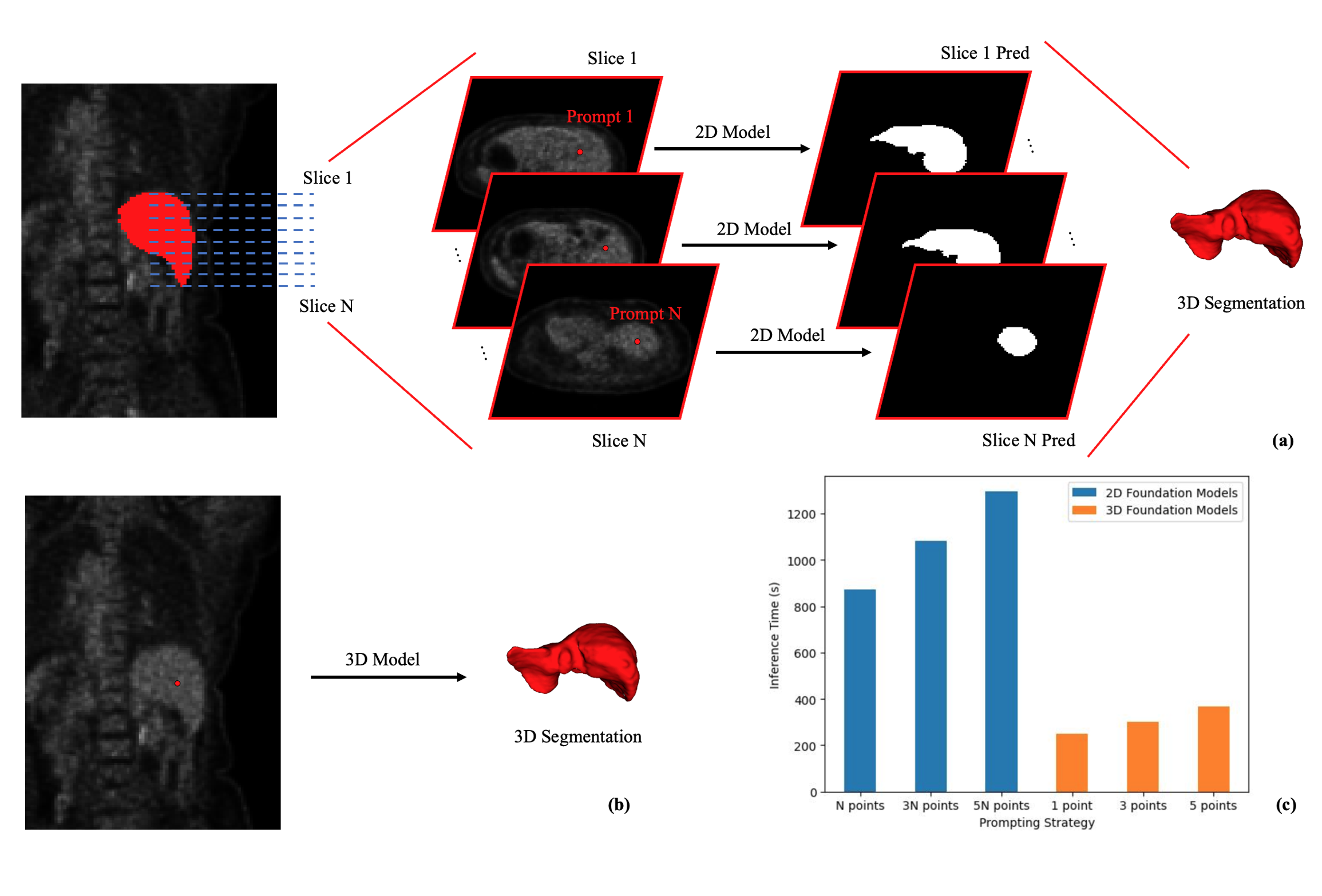}
	\caption{Visualization of different prompting strategies. (a) The segmentation workflow of 2D foundation models. (b) The segmentation workflow of 3D foundation models. (c) Inference time comparison of different prompting strategies.}
    \label{prompting}
\end{figure*}

As stated in the article, the input manual prompts are simulated based on the ground-truth mask for interactive segmentation.
Since the original SAM \cite{SAM} and MedSAM \cite{MedSAM} are designed for 2D segmentation tasks and cannot handle 3D inputs directly, a slice-by-slice procedure is conducted for the segmentation of the volume. The segmentation procedure of 2D foundation models necessitate input prompts for each 2D slice containing the target.
In contrast, SegAnyPET and other 3D SAM medical adaptions \cite{SAM-Med3D} can be directly utilized to segment the target organs from input volume with one or a few prompts. 
\Cref{prompting} (a) and (b) present the visualization of the segmentation workflow of 2D and 3D foundation models. Based on the comparison in \cref{prompting} (c), directly utilizing 3D foundation model for promptable segmentation can reduce the need of manual prompting with less inference time.

\subsection*{B.4. Evaluation Metric}

We use the Dice Similarity Coefficient (DSC) as the evaluation metric of the segmentation task, which is a widely used metric in the field of image segmentation to evaluate the similarity between two sets. The formula for DSC is given by:

\begin{equation*}
DSC(G, S) = \frac{2|G\cap S|}{|G| + |S|}
\label{DSC}
\end{equation*}
where $G$ represents the ground truth segmentation and $S$ represents the predicted segmentation. The DSC is calculated by taking twice the size of the intersection and dividing it by the sum of the sizes of the two sets. This normalization ensures that the coefficient ranges from 0 to 1, where 1 indicates perfect overlap between the ground truth and the predicted segmentation.

\section*{C. Additional Experiments and Discussion}

\subsection*{C.1. Generalization to Tumor Segmentation}

\begin{table}[h]
    \centering
    \normalsize
    \setlength\tabcolsep{9pt}
	\renewcommand\arraystretch{1.08}
	\begin{tabular}{c|c|c}
		\hline 	\hline
		Strategy  & Prompts & Tumor DSC \\ \hline
\multirow{3}{*}{SAM \cite{SAM}}  & N points & 19.03 \\
& 3N points & 27.89 \\
& 5N points & 40.91 \\ \hline 
\multirow{3}{*}{MedSAM \cite{MedSAM}} & N points & 1.77 \\
& 3N points & 26.13 \\
& 5N points & 32.14 \\ \hline 
\multirow{3}{*}{SAM-Med3D \cite{SAM-Med3D}} & 1 point & 11.45   \\ 
& 3 points & 14.93 \\
& 5 points & 16.32 \\  \hline 
\multirow{3}{*}{SegAnyPET} & 1 point & 19.38   \\ 
& 3 points & 24.57 \\
& 5 points & 24.93 \\  \hline \hline
	\end{tabular}
	\caption{Generalization performance to unseen out-of-distribution AutoPET dataset for zero-shot interactive tumor segmentation with comparison to state-of-the-art segmentation foundation models for zero-shot interactive segmentation from PET images.} \label{Table_Tumor}
\end{table}

\begin{figure*}[t]
    \centering
	\includegraphics[width=\linewidth]{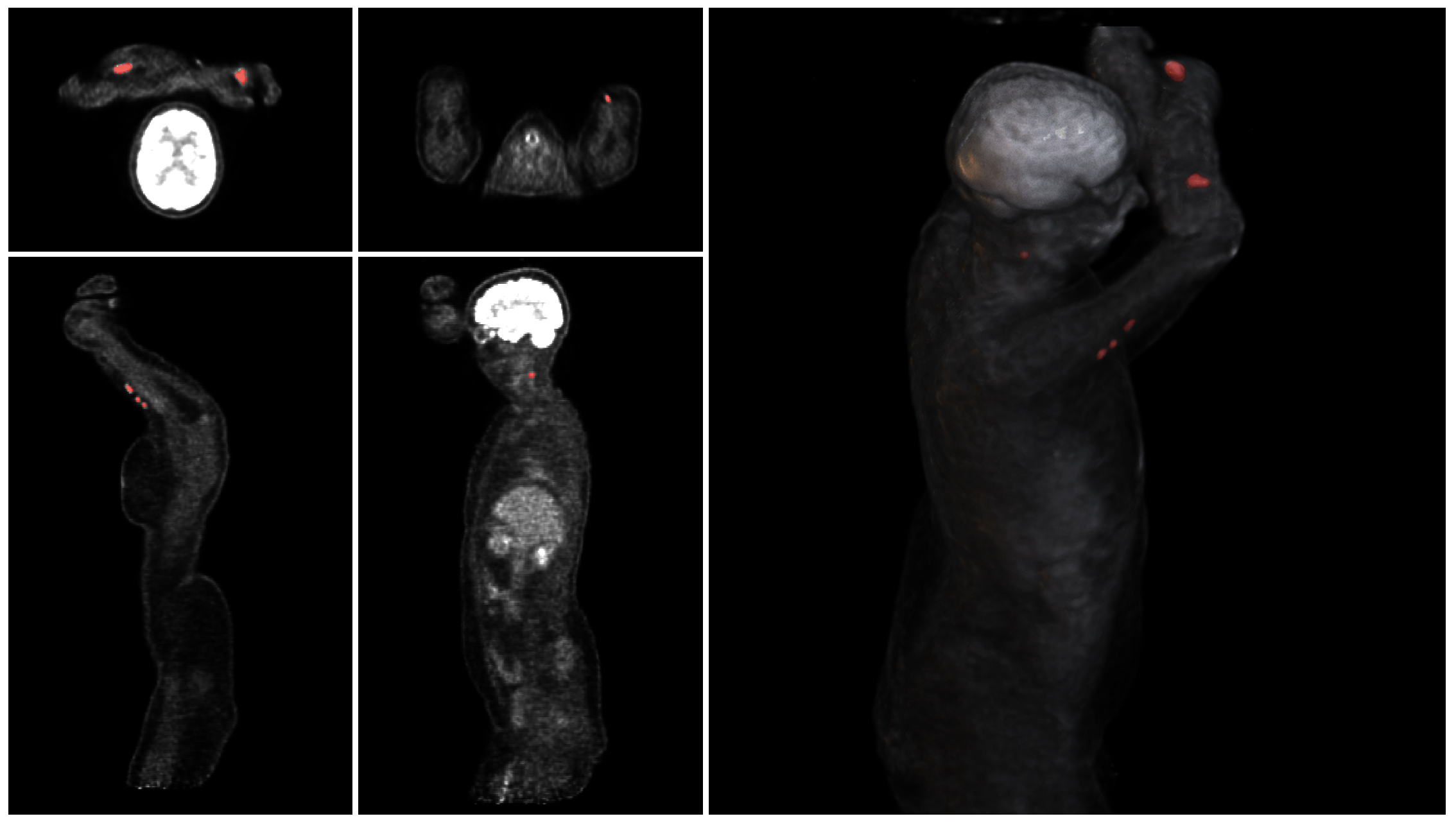}
	\caption{Visualization of an example case for whole-body tumor lesion segmentation of AutoPET dataset. The tumor regions are visualized in red.}
    \label{tumor_vis}
\end{figure*}

In addition to the internal and external evaluation on organ segmentation, we also compare SegAnyPET with other state-of-the-art segmentation foundation models for zero-shot tumor segmentation on AutoPET dataset.
\Cref{Table_Tumor} presents the experimental results under different prompt settings. Contrary to the conclusions of organ segmentation, we observe that slice-by-slice segmentation of 2D foundation models outperforms 3D foundation models.
A significant difference is that the target organs in our task are all continuous entities, while the whole-body tumors in AutoPET dataset are scattered multiple small targets located in various places, as shown in \cref{tumor_vis}. Therefore, the 3D model cannot directly segment all these scattered tumors out using one or a few prompt points, which also indicates the limitations of our method for this task.
As an important application scenario, we aim to enlarge the dataset with instance-level tumor annotation for training and evaluation of tumor lesion segmentation in the future.

\begin{table*}[h]
\centering
\setlength\tabcolsep{4pt}
\renewcommand\arraystretch{1.4}
\begin{tabular}{c|c|ccc|ccc|ccc}
\hline
Model & TotalSegmentator \cite{wasserthal2023totalsegmentator} & \multicolumn{3}{c}{SAM-Med3D-turbo} & \multicolumn{3}{|c}{SAM-Med3D-turbo} & \multicolumn{3}{|c}{SegAnyPET} \\ \hline
Modality & Registrated CT & \multicolumn{3}{c|}{Registrated CT} & \multicolumn{3}{c|}{PET} & \multicolumn{3}{c}{PET} \\ \hline
Prompt & Auto & 1 point & 3 points & 5 points & 1 point & 3 points & 5 points & 1 point & 3 points & 5 points \\ \hline
Avg DSC & 88.71 & 66.59 & 73.77 & 76.01 & 72.09 & 76.58 & 78.35 & 90.49 & 90.90 & 91.05\\
\hline
\end{tabular}
	\caption{Quantitative comparison different automatic and promptablt segmentation models for organ segmentation from CT and PET images.} \label{Table_CT}
\end{table*}

% \begin{table}[t]
%     \centering
%     \normalsize
%     \setlength\tabcolsep{4pt}
% 	\renewcommand\arraystretch{1.2}
% 	\begin{tabular}{c|c|c|c}
% 		\hline 	\hline
% 		Model & Modality  & Prompts & Avg DSC \\ \hline
%         TotalSegmentator & Registrated CT & Auto & 88.71  \\ \hline
% \multirow{3}{*}{SAM-Med3D-turbo} & & 1 point & 66.59   \\ 
% & Registrated CT & 3 points & 73.77 \\
% && 5 points & 76.01 \\  \hline 
% \multirow{3}{*}{SAM-Med3D-turbo} & & 1 point & 72.09   \\ 
% & PET & 3 points & 76.58 \\
% && 5 points & 78.35 \\  \hline 
% \multirow{3}{*}{SegAnyPET} & & 1 point & 90.49   \\ 
% &PET& 3 points & 90.90 \\
% && 5 points & 91.05 \\  \hline \hline
% 	\end{tabular}
% 	\caption{Quantitative comparison of organ segmentation models from CT and PET images.} \label{Table_CT}
% \end{table}

\subsection*{C.2. Comparison with CT Segmentation}

Since the data used in this work are whole-body PET/CT images, we conduct additional evaluations with models for automatic and promptable organ segmentation from CT images.
Given that the segmentation is used to evaluate the organ metabolic intensity from PET, it is necessary to register the CT to the resolution of PET before segmentation. Through the experimental comparison between the registered CT and PET in \cref{Table_CT}, we observe the performance of segmentation from CT is inferior to that of direct segmentation from PET.
Furthermore, developing a PET segmentation model can be compatible to different scenarios such as PET/MRI or CT-free PET with self-attenuation correction.

% {
%     \small
%     \bibliographystyle{ieeenat_fullname}
%     \bibliography{main}
% }

\end{document}